\documentclass{esannV2}
\usepackage[dvips]{graphicx}
\usepackage[latin1]{inputenc}
\usepackage{amssymb,amsmath,array,amsfonts}
\usepackage{graphics} 
\usepackage{color}
\usepackage{booktabs}
\usepackage{subfigure}
\usepackage{multirow}
\usepackage{hyperref}

\usepackage[style=base]{caption}  %% `skip` here does not seem to have any effect 
\DeclareCaptionFormat{side}{\hspace*{0.2cm}\parbox{3.9cm}{#1#2#3}}
\DeclareCaptionFormat{side2}{\hspace*{-0.5cm}\parbox{3cm}{#1#2#3}}
\DeclareCaptionFormat{centered}{\hspace*{2.3cm}\parbox{12cm}{#1#2#3}}
\DeclareCaptionFormat{centered2}{\hspace*{0.1cm}\parbox{11.5cm}{#1#2#3}}
\usepackage[rightcaption]{sidecap}
\sidecaptionvpos{figure}{c}

\definecolor{darkgreen}{rgb}{0,0.6,0.2}

\begin{document}
%style file for ESANN manuscripts
\title{Bidirectional deep-readout echo state networks}

%***********************************************************************
% AUTHORS INFORMATION AREA
%***********************************************************************
\author{Filippo M. Bianchi$^1$, Simone Scardapane$^2$, Sigurd L{\o}kse$^1$ and Robert Jenssen$^1$
%
% Optional short acknowledgment: remove next line if non-needed
\thanks{This work is partially funded by the Norwegian Research Council FRIPRO grant no. 239844 on the \emph{Next Generation Learning Machines} and IKTPLUSS grant no. 270738 \emph{Deep Learning for Health}.}
%
% DO NOT MODIFY THE FOLLOWING '\vspace' ARGUMENT
\vspace{.3cm}\\
%
% Addresses and institutions (remove "1- " in case of a single institution)
1- Machine Learning Group -- UiT the Arctic University of Norway \\
%
% Remove the next three lines in case of a single institution
\vspace{.1cm}\\
2- Sapienza University of Rome -- Dept. of Information Engineering, \\
Electronics and Telecommunications (DIET) \\
}
%***********************************************************************
% END OF AUTHORS INFORMATION AREA
%***********************************************************************

\maketitle

\begin{abstract}
We propose a deep architecture for the classification of multivariate time series.
By means of a recurrent and untrained reservoir we generate a vectorial representation that embeds temporal relationships in the data.
To improve the memorization capability, we implement a bidirectional reservoir, whose last state captures also past dependencies in the input.
We apply dimensionality reduction to the final reservoir states to obtain compressed fixed size representations of the time series.
These are subsequently fed into a deep feedforward network trained to perform the final classification.
We test our architecture on benchmark datasets and on a real-world use-case of blood samples classification.
Results show that our method performs better than a standard echo state network and, at the same time, achieves results comparable to a fully-trained recurrent network, but with a faster training.
\end{abstract}

\section{Introduction}

Reservoir computing (RC) is an established paradigm for modeling nonlinear temporal sequences~\cite{lukovsevivcius2009reservoir}. 
In machine learning tasks, echo state networks (ESNs) are the most common RC models, wherein the input sequence is projected to a high-dimensional space through the use of a (fixed) nonlinear recurrent reservoir~\cite{lukovsevivcius2009reservoir}. 
Learning is performed by applying simple linear techniques in the high-dimensional reservoir space. 
The lack of flexibility in the recurrent part is balanced by a range of advantages, including faster training compared to other recurrent neural networks (RNNs).
In tasks requiring a limited amount of temporal memory, ESNs achieve state-of-the-art results in many real-world scenarios constrained by time budgets, low-power hardware and limited data~\cite{scardapane2017randomness}. 
On the other hand, fully-trained RNNs trade architectural and training complexity with more accurate representations and a larger memory capability~\cite{NIPS2014_5346}.

We propose an architecture for time series classification called Bidirectional Deep-readout ESN (BDESN), which combines the training speed of RC with the accuracy of trainable RNNs.
We equip our model with a \textit{bidirectional reservoir} to curtail the short-term memory limitation in ESNs.
Bidirectional architectures have been successfully applied in RNNs to extract temporal features from the time series that accounts also for dependencies very far in time~\cite{graves2005framewise}.
Differently from the classic linear readout found in ESNs, we generate the desired output by processing the temporal features with a deep feedforward network trained with gradient descent. 
Although we loose a closed form solution, the whole architecture can still be trained quickly, since the recurrent part is fixed and the time consuming unfolding procedure~\cite{bianchi2017recurrent} is avoided. 
Additionally, we can leverage recent advancements in deep learning to speed up the learning~\cite{goodfellow2016deep}. 
The temporal features generated by the bidirecitonal reservoir are projected into a subspace before being fed into to feedforward network, to reduce model and training complexity~\cite{lokse2017training}.
%For this task, we apply a dimensionality reduction procedure, which is an effective regularizer in RC architectures.
Deep feedforward networks on top of RNNs demonstrated to enhance the representative power of the state space of a recurrent model~\cite{pascanu2013construct}.
However, so far their application in reservoir computing has been limited, due to difficulties in training~\cite{lukovsevivcius2009reservoir}.
We note that our model differs from the recently proposed deep ESNs \cite{gallicchio2016deep, gallicchio2017deep}, which are built by concatenating multiple (fixed) reservoirs, with no modification of the adaptable part. 
The two ideas are in fact complementary, and can in principle be used together. 
%Nonlinear techniques for training the output layer were not found to provide improvements in performance in the original ESN implementations \cite{lukovsevivcius2009reservoir}, except in some specific situations like chaotic series forecasting \cite{shi2007support}.

We empirically validate our model against standard ESNs and fully-trainable RNNs on multiple benchmark datasets. 
We show that BDESN vastly outperforms an ESN, achieving a competitive accuracy with respect to RNNs, while being orders of magnitude faster to train. 
%(even after all hyperparameters have been carefully fine-tuned)

\section{Methodology}

\begin{figure}[!ht]
\centering
\includegraphics[width=0.75\columnwidth, keepaspectratio,trim={0cm 0.1cm 0cm 0cm},clip]{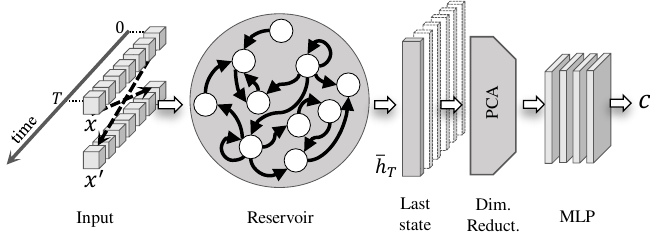}
\caption{Schematic illustration of the BDESN architecture.}
	\label{fig:BDESN_arch}
\end{figure}

The BDESN assigns to a multivariate time series (MTS) $\boldsymbol{x} = \{\mathbf{x}_t\}_{t=0}^T$ a class label $\boldsymbol{c}$ through a three-step procedure.
First, a special recurrent \emph{reservoir} generates a representation of $\boldsymbol{x}$ that has fixed (large) size and embeds temporal features.
Then, a \emph{dimensionality reduction algorithm} projects the reservoir outputs in a lower dimensional space.
Finally, a \emph{multilayer perceptron} (MLP) classifies the vectorial representation of $\boldsymbol{x}$.
In Fig.~\ref{fig:BDESN_arch} we depict the whole architecture, whose details are discussed in the following.

\captionsetup{format=centered,font=footnotesize,labelfont=bf,labelsep=period}

\subsection{Extracting features with a reservoir}
The reservoir is governed by the following state-update equation
\begin{equation}
 \label{eq:reservoir}
 \mathbf{h}_t = f(\mathbf{W}^h \mathbf{h}_{t-1} + \mathbf{W}^i \mathbf{x}_{t}),
\end{equation}
where $\mathbf{h}_t$ is the internal time-dependent state, which combines the current input $\mathbf{x}_t$ and the previous computations $\mathbf{h}_{t-1}$.
The function $f$ is a nonlinear activation (usually a tanh), $\mathbf{W}^h$ is a sparse matrix that defines the recurrent self-connections in the reservoir, and $\mathbf{W}^i$ defines input connections.
Both matrices are randomly generated and left untrained, and the reservoir behavior is controlled mainly by three hyperparameters.
These are the state size $N$, the spectral radius $\rho$ of $\mathbf{W}^h$, and scaling of the inputs $\omega$.
Through an optimal tuning of these hyperparameters the reservoir produces rich dynamics and its internal states can be used to solve many prediction and classification tasks~\cite{lukovsevivcius2009reservoir}.

% vanishing memory and bidirectional reservoir
The last state $\mathbf{h}_T$ generated by the reservoir, after the whole input $\boldsymbol{x}$ is processed, is a high-level representation of fixed size that embeds the temporal dependencies of $\boldsymbol{x}$.
Since the reservoir trades its internal stability with a vanishing memory of the past inputs~\cite{7765110}, at time $T$ the state maintains scarce information about the first inputs.
To alleviate this issue, we feed to the same reservoir also the MTS in reverse order, $\boldsymbol{x'} = \{\mathbf{x}_{T-t}\}_{t=0}^T$ and we generate a new representation $\mathbf{h'}_T$ that is more influenced by the first inputs.
A final representation is obtained by concatenating the two states, $\bar{\mathbf{h}}_T = \left[ \mathbf{h}_T; \mathbf{h'}_T \right]^T$.
A bidirectional reservoir has been recently used for time series prediction~\cite{rodan2017bidirectional}.
However, the architecture proposed in this work is significantly different, mainly because we deal with classification tasks.

\subsection{Dimensionality Reduction}
A dimensionality reduction method maps data from high to a lower dimensional space, usually relying on unsupervised criteria.
In a classification setting supervised techniques can also be adopted to improve the dimensionality reduction procedure, although we do not consider them in this paper.

Popular algorithms are principal component analysis (PCA) and kernel PCA, which project data on the first $d$ eigenvectors of a covariance matrix.
PCA and kPCA have been already successfully applied to ESN reservoir states, improving its prediction and modeling capabilities~\cite{lokse2017training}.
In BDESN, dimensionality reduction provides a strong regularization that prevents overfitting in the classifier operating on the reservoir states.

\subsection{Multilayer Perceptron}

In a standard ESN, the output layer is a linear readout that is quickly trained by solving a convex optimization problem. However, we hypothesize that a simple linear model does not possess sufficient representational power for modeling the high-level embeddings resulting from our reservoir space.
For this reason, in BDESN the classification is performed by a MLP with $L$ layers on the representations generated at the previous steps. 
The number of layers in the MLP thus determines a ``feedforward'' depth to the RNN \cite{pascanu2013construct}. 
Deep MLPs are known for their capability of disentangling factors of variations from high-dimensional spaces, and can nowadays be trained efficiently with the use of sophisticated regularization techniques. In particular, we leverage two  procedures, namely dropout and the minimization of the $\mathrm{L}_2$ norm of the model parameters (i.e., weight decay).
% MLP is an universal function approximator that can learn complex representations of the input by stacking more layers of neurons configured with non-linear activations.
Thanks to the dimensionality reduction step, we can greatly reduce the number of parameters in the MLP, hence its variance, speed up the computation and obtain a faster convergence during training.

\section{Experiments}

To evaluate the proposed architecture, we compare the classification accuracy and the training time of BDESN with two different RNNs. 
The first is a standard ESN, configured with the same reservoir (only unidirectional) of BDESN and with linear readout trained with ridge regression.
The second is a network where the vectorial representations are generated by a recurrent layer of gated recurrent units (GRUs), which are fed into a MLP equivalent to the one of BDESN.
Contrarily to the reservoir, the GRU layer is trained supervisedly.

Each RNN depends on different hyperparameters, which are optimized by means of random search.
The optimized RNNs are trained 10 times, using random and independent model initializations on a Intel Core i7-6850K CPU@4GHz.
The code is implemented in Tensorflow and is publicly available online, along with a detailed description of the experimental setup and validation procedure\footnote{\url{github.com/FilippoMB/Bidirectional-Deep-reservoir-Echo-State-Network}}.

\subsection{Benchmark datasets}
We perform classification of MTS from different datasets, whose details are reported in Tab.~\ref{tab:dataset}.

% :::::::::::::::::::::::::: TAB DESCRIPTION ::::::::::::::::::::::::::
\captionsetup{format=side,font=footnotesize,labelfont=bf,labelsep=period}
\bgroup
\def\arraystretch{0.9} %vertical padding
\setlength\tabcolsep{.1em} %horizontal padding
\begin{SCtable}[1.2][!ht]
\small
\centering
\caption{Time series dataset details. Column 2 to 5 report the number of variables (\#$V$), samples in training and test set, and number of classes (\#$C$), respectively. 
$T_{min}$ is the length of the shortest MTS in the dataset and $T_{max}$ the longest MTS.}
\label{tab:dataset}
\begin{tabular}{@{}ll@{ }@{ }l@{ }@{ }@{ }l@{ }@{ }@{ }ll@{ }l@{ }lc@{}}
\cmidrule[1.5pt]{1-8}
\textbf{Dataset} & \#$\boldsymbol{V}$ & \textbf{Train} & \textbf{Test} & \#$\boldsymbol{C}$ & $\boldsymbol{T}_{min}$ & $\boldsymbol{T}_{max}$ & \textbf{Source} \\
\cmidrule[.5pt]{1-8}
DistPhal & 1 & 400 & 139 & 3 & 80 & 80 & UCR \\
ECG & 2 & 100 & 100 & 2 & 39 & 152 & UCR \\
Libras & 2 & 180 & 180 & 15 & 45 & 45 & UCI \\
Ch.Traj. & 3 & 300 & 2558 & 20 & 109 & 205 & UCI \\
Wafer & 6 & 298 & 896 & 2 & 104 & 198 & UCR \\
Jp.Vow. & 12 & 270 & 370 & 9 & 7 & 29 & UCI \\
%Arab. Dig. & 13 & 6600 & 2200 & 10 & 4 & 93 & UCI \\
%Auslan & 22 & 1140 & 1425 & 95 & 45 & 136 & UCI \\
\cmidrule[1.5pt]{1-8}
\end{tabular}
\end{SCtable}
\egroup
% :::::::::::::::::::::::::::::::::::::::::::::::::::::::::::::::::::::

Classification accuracy on the test sets and training time (in minutes) obtained by each RNN are reported in Tab.~\ref{tab:results}.
%
% :::::::::::::::::::::::::: TAB RESULTS ::::::::::::::::::::::::::
\captionsetup{format=centered2,font=footnotesize,labelfont=bf,labelsep=period}
\bgroup
\def\arraystretch{0.9} %vertical padding
\setlength\tabcolsep{.4em} %horizontal padding
\begin{table}[!ht]
\small
\centering
\caption{Classification accuracy and training time (minutes) obtained by each RNN on different datasets in $10$ independent runs. Best average accuracies are in bold.}
\label{tab:results}
\begin{tabular}{l|cc|cc|cc}
\cmidrule[1.5pt]{1-7}
\multicolumn{1}{c}{\multirow{ 2}{*}{\textbf{Dataset}}} &  \multicolumn{2}{c}{\textbf{ESN}} & \multicolumn{2}{c}{\textbf{GRU}} & \multicolumn{2}{c}{\textbf{BDESN}} \\
\multicolumn{1}{c}{} & \multicolumn{1}{c}{\textbf{Accuracy}} & \multicolumn{1}{c}{\textbf{Time}} & \multicolumn{1}{c}{\textbf{Accuracy}} & \multicolumn{1}{c}{\textbf{Time}} & \multicolumn{1}{c}{\textbf{Accuracy}} & \multicolumn{1}{c}{\textbf{Time}} \\
\cmidrule[.5pt]{1-7}
DistPhal & 71.5$\pm$3.3 & 0.11 & 69.4$\pm$2.4 & 34.33 & \textbf{73.7}$\pm$0.91 & 1.15\\
ECG & 65.6$\pm$4.63 & 0.03 & \textbf{79.1}$\pm$4.3 & 10.35 & 78.5$\pm$1.5 & 0.22 \\
Libras & 76.2$\pm$2.77 & 0.03 & 72.7$\pm$4.8 & 5.27 & \textbf{78.4}$\pm$1.9 & 0.53 \\
Ch.Traj. & 43.4$\pm$2.72 & 0.23 & 72.1$\pm$6.4 & 40.28 & \textbf{77.6}$\pm$5.53 & 0.63\\
Wafer & 89.7$\pm$0.9 & 0.06 & \textbf{98.2}$\pm$0.9 & 35.38 & 90.3$\pm$0.8 & 0.76 \\
Jp. Vow. & 85.8$\pm$2.34 & 0.06 & 94.45$\pm$1.88 & 6.03 & \textbf{94.72}$\pm$1.12 & 0.58 \\
%Arab. Dig. &  &  &  &  &  & \\
%Auslan &  &  &  &  &  & \\
\cmidrule[1.5pt]{1-7}
\end{tabular}
\end{table}
\egroup
% :::::::::::::::::::::::::::::::::::::::::::::::::::::::::::::::::::::
%
\captionsetup{format=side2,font=footnotesize,labelfont=bf,labelsep=period}
\begin{SCfigure}[1.2][!ht]
\centering
\includegraphics[width=0.6\columnwidth, keepaspectratio,trim={0cm 0.25cm 0cm 0cm},clip]{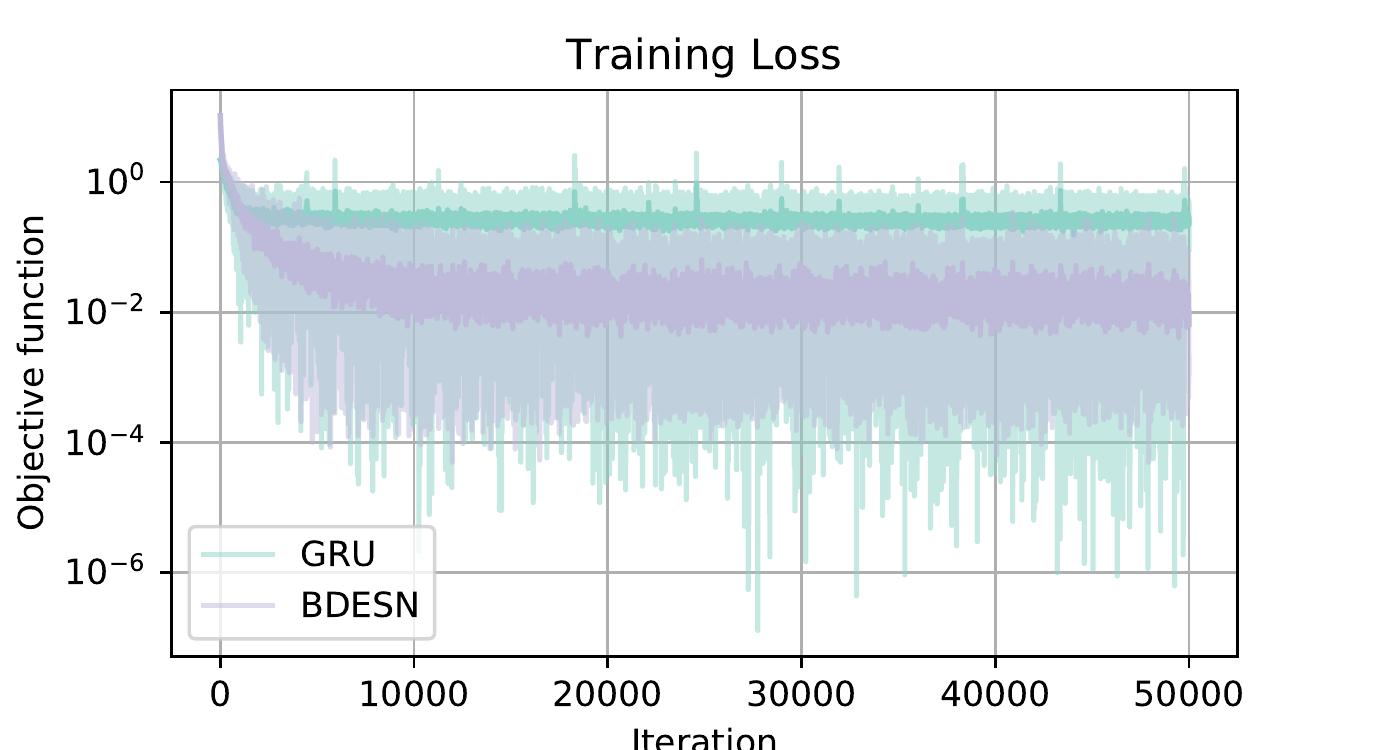}
\caption{Cross entropy loss assumed by GRU and BDESN during training. In BDESN, the loss is more stable and, in average, assumes lower values.}
	\label{fig:train_loss}
\end{SCfigure}

Results show that BDESN consistently performs better than ESN, by obtaining an increment of the accuracy up to 30 percentage points (Character Trajectory dataset).
BDESN outperforms GRU, sometimes slightly, in several classification tasks, except in the Wafer dataset.
Average GRU accuracy is higher also in ECG, but the difference is not significant due to the high standard deviation in the results.
While BDESN is slower than ESN, it can still be trained up to 70 times faster than GRU, providing a huge impact in terms of the required computational resources.

Even if GRU during the training phase may reach a lower cross entropy on the training set, in BDESN the average loss is usually lower and the training is more stable.
For example, in Fig.~\ref{fig:train_loss} we report the cross entropy values reached during training on the Japanese Vowels dataset. We observe that GRU is more unstable and converges to less accurate values than BDESN.

\subsection{Classification of blood samples time series}

We test our method also on a real-world dataset of blood measurements collected from patients undergoing a gastrointestinal surgery at University Hospital of North Norway in 2004--2012. 
Each patient in the dataset is represented by a MTS of $10$ blood samples measurements extracted within $20$ days after surgery.
We consider the problem of classifying patients with and without surgical site infections from their blood samples.
The MTS contains missing data, corresponding to measurements not collected for a given patient in one day of the observation period, which are replaced with mean-imputation.
The dataset consists of $883$ MTS, of which $232$ are relative to patients with infections. 
%Data are in random order and the first $80\%$ are used as training set and the rest as test set.

% :::::::::::::::::::::::::: TAB BLOOD ::::::::::::::::::::::::::
\captionsetup{format=side,font=footnotesize,labelfont=bf,labelsep=period}
\bgroup
\def\arraystretch{0.95} %vertical padding
\setlength\tabcolsep{.3em} %horizontal padding
\begin{SCtable}[1][!ht]
\small
\centering
\caption{F1 score and time of the different models on the blood samples dataset. Best average result is in bold.}
\label{tab:blood}
\begin{tabular}{l|ccc}
\cmidrule[1.5pt]{1-3}
\textbf{RNN} & \textbf{F1 score} & \textbf{Time (mins)} \\
\cmidrule[.5pt]{1-3}
ESN & 0.734$\pm$0.041 & 0.01 \\
GRU & \textbf{0.779}$\pm$0.018 & 10.54\\
BDESN & 0.761$\pm$0.054 & 1.87 \\
\cmidrule[1.5pt]{1-3}
\end{tabular}
\end{SCtable}
\egroup
% :::::::::::::::::::::::::::::::::::::::::::::::::::::::::::::::::::::

In Tab.~\ref{tab:blood} we report the F1 score and time obtained by the different models on the blood samples data.
Since the dataset is imbalanced, F1 score in this case is a more suitable measure of performance than accuracy.
We observe that the performance of GRU and BDESN are comparable and both are better than ESN.
GRU mean F1 score is somehow higher and, by looking at the standard deviation, we notice its performance to be slightly more stable.
However, GRU trades this moderate improvement with a significant increment in training time, which is 1 and 2 orders of magnitude higher than BDESN and ESN, respectively.

\section{Conclusions}

We proposed a novel recurrent architecture for time series classification, the bidirectional deep-readout ESN, which captures dependencies in the data forward and backward in time, by means of a large untrained reservoir.
The representation yielded by the reservoir is firstly regularized with PCA and then classified by a MLP trained supervisedly.
Our approach trains much faster than a RNN configured with GRU cells to generate the same kind of representations.
We tested our architecture on several benchmark datasets and one real-world use-case of time series classification.
Results indicate that BDESN performance is much better than ESN and sometime even better than a fully trained GRU network.

% ****************************************************************************
% BIBLIOGRAPHY AREA
% ****************************************************************************

\begin{footnotesize}

\bibliographystyle{unsrt}
\bibliography{refs}

\end{footnotesize}

% ****************************************************************************
% END OF BIBLIOGRAPHY AREA
% ****************************************************************************

\end{document}